\documentclass[letterpaper]{article}
\usepackage{aaai24}
\usepackage{times} 
\usepackage{helvet} 
\usepackage{courier} 
\usepackage[hyphens]{url} 
\usepackage{graphicx}
\usepackage{natbib}
\usepackage{caption} 
\usepackage{algorithm}
\usepackage{algorithmic}

\newcommand{\etal}{\textit{et~al.\,}}
\newcommand{\ie}{\textit{i.e.}}

\usepackage{float}
\usepackage{subfigure}
\usepackage{times}
\usepackage{helvet}
\usepackage{courier}
\usepackage{multirow}
\usepackage{amsmath,amssymb,amsfonts}
\usepackage{color}
\usepackage{comment}
\usepackage{amstext}
\usepackage[dvipsnames]{xcolor}
\usepackage{comment}
\usepackage{enumitem}
\usepackage{diagbox}

\usepackage{newfloat}
\usepackage{listings}
\DeclareCaptionStyle{ruled}{labelfont=normalfont,labelsep=colon,strut=off}
\floatstyle{ruled}
\newfloat{listing}{tb}{lst}{}
\floatname{listing}{Listing}
\title{ROG$_{PL}$: Robust Open-Set Graph Learning via Region-Based Prototype Learning}
\author{
    Qin Zhang\textsuperscript{\rm 1},
    Xiaowei Li\textsuperscript{\rm 1},
    Jiexin Lu\textsuperscript{\rm 1},
    Liping Qiu\textsuperscript{\rm 1},
    Shirui Pan\textsuperscript{\rm 2},
    Xiaojun Chen\textsuperscript{\rm 1}\footnote{Corresponding author},
    Junyang Chen\textsuperscript{\rm 1}
}
\affiliations{
    \textsuperscript{\rm 1}College of Computer Science and Software Engineering, Shenzhen University, China
    \textsuperscript{\rm 2}School of Information and Communication Technology, Griffith University, Australia.\\
    qinzhang@szu.edu.cn, \{lixiaowei2022, lujiexin2022, qiuliping2021\}@email.szu.edu.cn, s.pan@griffith.edu.au, \\ \{xjchen, junyangchen\}@szu.edu.cn
}

\begin{document}
\maketitle
\begin{abstract}
Open-set graph learning is a practical task that aims to classify the known class nodes and to identify unknown class samples as unknowns.
Conventional node classification methods usually perform unsatisfactorily in open-set scenarios due to the complex data they encounter, such as out-of-distribution (OOD) data and in-distribution (IND) noise. 
OOD data are samples that do not belong to any known classes.
They are outliers if they occur in training (OOD noise), and open-set samples if they occur in testing. 
IND noise are  training samples which are assigned incorrect labels. 
The existence of IND noise and OOD noise is prevalent, which usually cause the ambiguity problem, including the \textit{intra-class variety problem} and the \textit{inter-class confusion problem}.
Thus, to explore robust open-set learning methods is necessary and difficult, and it becomes even more difficult for non-IID graph data.
To this end, we propose a unified framework named ROG$_{PL}$ to achieve robust open-set learning on complex noisy graph data, by introducing prototype learning. 
In specific, ROG$_{PL}$ consists of two modules, \ie,  denoising via label propagation and open-set prototype learning via regions. 
The first module corrects noisy labels through similarity-based label propagation and removes low-confidence samples, to solve the intra-class variety problem caused by noise. 
The second module learns open-set prototypes for each known class via non-overlapped regions and remains both interior and border prototypes to remedy the inter-class confusion problem.
The two modules are iteratively updated under the constraints of  classification loss and prototype diversity loss. 
To the best of our knowledge, the proposed ROG$_{PL}$ is the first robust open-set node classification method for graph data with complex noise. 
Experimental evaluations of ROG$_{PL}$ on several benchmark graph datasets demonstrate that it has good performance.
\end{abstract}
\begin{figure}[htb]
\centering
\includegraphics[width=0.48\textwidth, height=9cm]{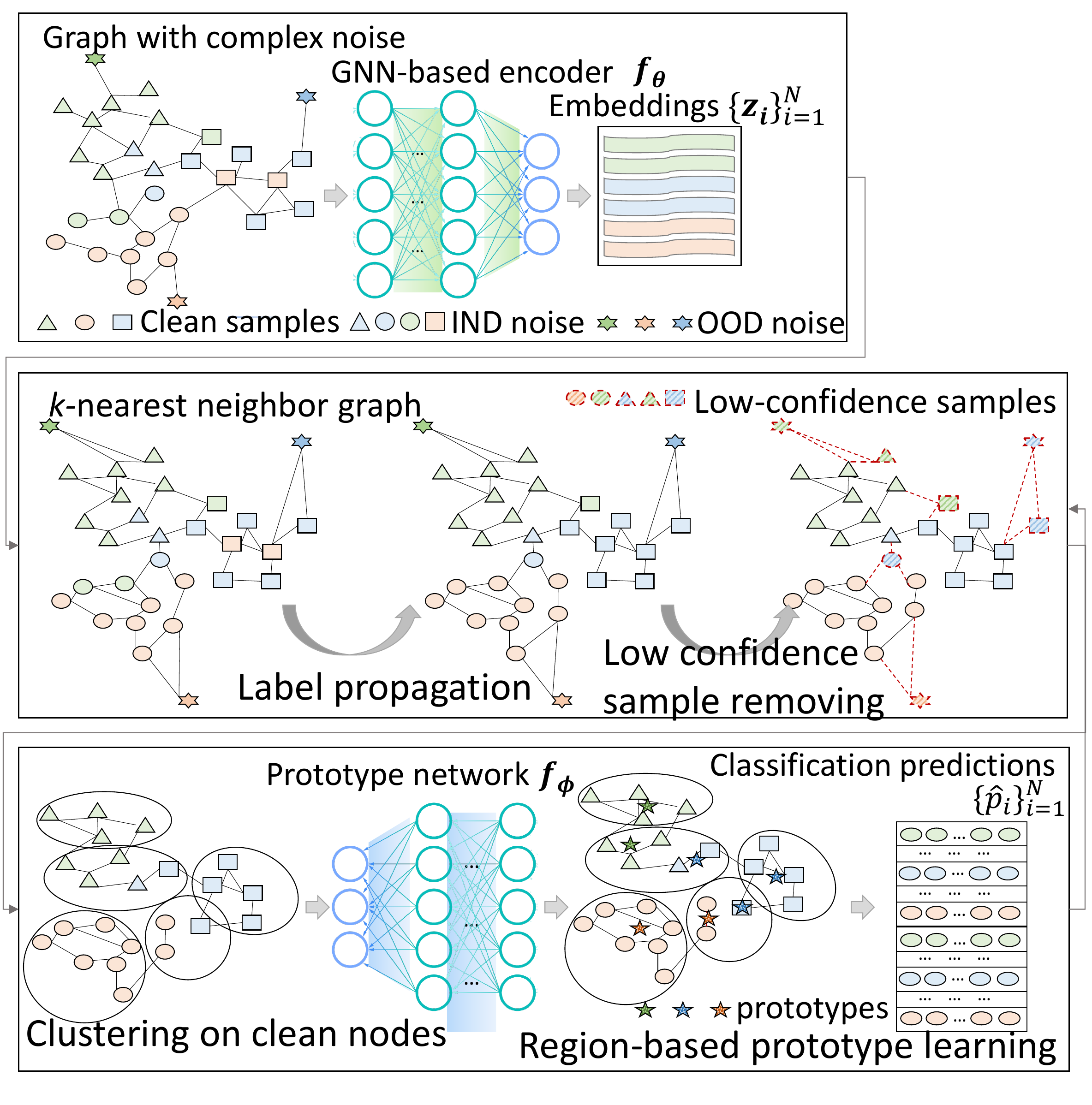} 
\caption{Overview of the proposed ROG$_{PL}$. In the latent representation space, two modules are designed: denoising via label propagation and open-set prototype learning via regions.
In specific, we first correct noisy labels through similarity-based label propagation and removes low-confidence samples, to solve the intra-class variety problem caused by noise. 
Then we learn open-set prototypes for each known class via non-overlapped regions and remains both interior and border prototypes to remedy the inter-class confusion problem. 
These two modules are iteratively updated under the constraints of classification loss and prototype diversity loss. } 
\label{Fig:framework}
\end{figure}
\section{Introduction}
Graph neural networks (GNNs) \cite{gilmer2017neural,guo2022orthogonal,hamilton2017inductive,tan2023ethereum} have become a prominent technique to analyze graph structured data in many real-world systems, such as traffic state prediction~\cite{zheng2020gman},  disease classification  \cite{chereda2019utilizing}, 
and user profile completion in  social networks~\cite{wong2021standing}.
The recent success of supervised GNNs is built upon two crucial cornerstones: that the training and test data are drawn from an \textit{identical distribution}, and that large-scale reliable \textit{ high-quality labeled } data are available for training. 
However, in real-world applications, large-scale labeled data drawn from the same distribution as test data are usually unavailable. 

Real-world applications normally are in open-set scenarios \cite{zhang-etal-2023-survey-efficient,nimah-etal-2021-protoinfomax-prototypical,zhang2022dynamic}, the existence of new-emerged out-of-distribution (OOD) samples, \ie  samples that do not belong to any known classes, is prevalent. 
They are outliers if they occur in training (OOD noise), and open-set samples if they occur in testing. 
Moreover, the manually generating clean labeled data set would involve domain experts evaluating the quality of collected data and thus is very expensive and time-consuming \cite{zhang2023denoising}. 
Alternatively, we can collect data and labels based on web search \cite{yu2018learning}, crowdsourcing \ \cite{fang2014active,li2017webvision} and user tags \cite{li2017learning,xiao2015learning}. These data and labels are cheap but inevitably noisy \cite{zhang2023denoising,li2022selective}. 

The presence of OOD noise and IND noise ( in-distribution samples with incorrect labels)  can be detrimental to GNNs~\cite{zhang2021understanding}, as they would cause the ambiguity problem, including the \textit{intra-class variety problem} in which samples with same class labels may contain objects of different semantic categories, and the \textit{inter-class confusion problem} in which samples of different intent classes may contain objects of similar semantic categories\cite{wang2023prototype}. The open-set unknown class samples occur during testing further complicates this problem. 

Thus, it is necessary to build robust open-set learning models 
\cite{wu2021ngc} that can learn from noisy data, achieving  classification of known class samples and  identification of  unknown open-set class samples during testing. 
Previous works mainly concentrated on robust learning or open-set learning separately, the problem of robust open-set learning with complex noise has not been sufficiently explored so far \cite{huang2022end,zhang2023g2pxy}, and it becomes even more difficult for non-IID graph data.

In this paper, as shown in Fig. \ref{Fig:framework}, for solving the problem of open-set graph learning with complex IND and OOD noise, we propose a new framework ROG$_{PL}$ to do robust open-set node classification.
It consists of two main steps:  denoising via label propagation and open-set prototype learning via regions.
Specifically, in the latent representation space, 
ROG$_{PL}$ first corrects noisy labels through similarity-based pseudo-label propagation and remove low-confidence samples, to solve the intra-class variety problem caused by noise. 
Then the second module learns open-set prototypes for each known class via non-overlapped regions and remains both interior and border prototypes to remedy the inter-class confusion problem.
The two modules are iteratively updated under the constraints of both classification loss and prototype diversity loss. 
Experimental evaluations of ROG$_{PL}$ on several benchmark graph datasets demonstrates that it has good performance.

\section{Preliminaries} 
This study focuses on the node classification problem for a graph.
A \textsl{graph} is denoted  as $G=(V,E,X)$, where $V=\{v_i | {i = 1,\ldots,N}\}$ is a set of $N$ nodes in the graph, and 
$E=\{e_{i,j} | i,j = 1,\ldots,N.$ $i\neq j \}$ is a set of  edges connecting pairs of nodes $v_i$ and $v_j$.
$X\in \mathbb{R}^{N\times s}$ denotes the feature matrix of  nodes, where $s$ is the dimension of node features. 
The feature vector of each node $v_i$ is indicated by $x_i \in X$. 
The topological structure of $G$ is represented by an adjacency matrix $A \in \mathbb{R}^{N \times N}$, where $A_{i,j}=1$ if the nodes $v_i$ and $v_j$ are connected,~\ie, $\exists e_{i,j} \in E$, and otherwise $A_{i,j}= 0$. 
The label matrix of $G$ is  $Y \in \mathbb{R}^{N \times C}$, where $C$ is the already-known node classes. 
If a label $c$ is assigned to a node $v_i \in V$, then $y_{i,c} = 1$, and otherwise, $y_{i,c} = 0$.

For a typical \textbf{closed-set node classification} problem, a GNN encoder $f_{\theta_g}$ takes the node features $X$ and adjacency matrix $A$ as input, aggregates the neighborhood information and outputs representations.
Then, a classifier $f_{\theta_c}$ is used to classify the nodes into $C$ already-known classes.
The GNN encoder and the classifier are optimized to minimize the expected risk ~\cite{yu2017open-holder38} in Eq. \eqref{eq:closedrisk}, assuming that the testing data $\mathcal{D}_{te}$ and the training data $\mathcal{D}_{tr}$ have the same feature space and label space, \ie,
\begin{equation}
    \label{eq:closedrisk}
    f^* = arg\min_{f\in \mathcal{H}} \mathbb{E}_{(x,y) \sim \mathcal{D}_{te}} \mathbb{I}(y\neq f(\theta_g, \theta_c; x,A))
\end{equation}
where $\mathcal{H}$ is the hypothesis space, $\mathbb{I}(\cdot)$ is the indicator function which outputs 1 if the expression holds and 0 otherwise. This function can be generally optimized using cross-entropy to distinguish between known classes.

In the \textbf{open-set node classification problem}, given a graph $G=(V,A,X)$, $\mathcal{D}_{tr}=(X,Y)$ denotes the training nodes.
The test nodes are denoted by $\overline{\mathcal{D}}_{te}=(X_{te},Y_{te})$, where  $X_{te}=S\cup U$, $Y_{te}= \{1,\ldots, C, C+1,\ldots\}$. The set $S$ is the nodes that belong to seen classes that already appeared in $\mathcal{D}_{tr}$ and $U$ is the set of nodes that do not belong to any seen class (\ie, unknown class nodes). 
The goal of open-set node classification is to learn a $(C+1)$-$class$ classifier $f_{\overline{\theta}_c}$ such that $f(\theta_g,\overline{\theta}_c; X_{te},\overline{A}): \{X_{te},\overline{A}\}\rightarrow \overline{\mathcal{Y}}$, $\overline{\mathcal{Y}}= \{1, \ldots , C, unknown\}$, by minimizing the expected risk \cite{yu2017open-holder38}: 
\begin{small}  
\begin{equation}
    \label{eq:openrisk}
    \overline{f}^* = arg\min_{f\in \mathcal{H}} \mathbb{E}_{(x,y) \sim \mathcal{\overline{D}}_{te}} \mathbb{I}(y\neq f(\theta_g, \overline{\theta}_c; x,\overline{A}))
\end{equation}
\end{small} 
where $\overline{A}$ is the adjacency matrix for $X_{te}$.
The predicted class $unknown\in\overline{\mathcal{Y}}$ consists of a group of novel categories, which may contain multiple classes. 
An intuitive way of transforming a close-set classifier into an open-set classifier is thresholding~\cite{hendrycks2016baseline-han7-holder10}.

In the problem of \textbf{Open-set node classification with IND noise and OOD noise}, given a graph $G=(V,A,X)$, $\mathcal{D}_{tr} = (X,Y=S)=\{x_i,y_i\}_{i=1}^N$ is the training set and $\mathcal{D}_{te} = (X_{te},Y_{te}=S\cup U)$ is the test set. 
In $\mathcal{D}_{tr}$, 
we assume that the instance-label pair $(x_i,y_i)$ , $1 \leq i \leq N$, consists of three types. 
Let $y_i^*$ denote the ground-truth label of $x_i$. A \textbf{clean sample} is a node whose assigned label matches the ground-truth label, i.e., $y_i=y_i^*$.
An \textbf{IND noise sample} is a node whose assigned label does not match the ground-truth label, but the node matches one of the classes in $S$, \ie $y_i\neq y_i^*, y_i^*\in S$.
An \textbf{OOD noise sample} is a node whose assigned label does not match the ground-truth label and any known class label neither, \ie $y_i \neq y_i^*, y_i^* \notin S$.
Moreover, under the strict setting of open-set classification,  it is assumed that the ground-truth label of OOD noise $y_i^* \notin U$, \ie, there is no overlap between the classes of OOD noise and the unknown classes in the test set.
Due to the memorization effect \cite{arpit2017closer}, noisy data can severely impair the performance of network training. Therefore, it is desirable to develop noise-robust methods for open-set node classification, which can handle complex and diverse noises.
The ultimate goal is to learn a noise-robust open-set node classifier that can minimize the expected risk in Eq. \eqref{eq:openrisk}.

\section{Methodology}  
For solving the problem of open-set graph learning with complex IND and OOD noise, we propose a new framework ROG$_{PL}$, which consists of two main steps:  denoising via label propagation and open-set prototype learning via regions.
As shown in Fig. \ref{Fig:framework}, 
in the latent representation space, 
we first corrects noisy labels through soft pseudo-label propagation and removes low-confidence samples, to solve the intra-class variety problem caused by noise. 
Then we learn open-set prototypes for each known class via non-overlapped regions and remains both interior and border prototypes to remedy the inter-class confusion problem.
The two modules are iteratively updated under the constraints of both classification loss and prototype diversity loss. 

\subsection{Denoising via Label Propagation}
Inspired by label propagation algorithms for semi-supervised learning\cite{grandvalet2004semi,iscen2018mining,chandra2016fast}, which seek to transfer labels from supervised examples to neighboring unsupervised examples according to their similarity in feature space, we leverage neighbour consistency to modify the supervision of each sample and to correct noise.
First, we build a $k$-nearest neighbor graph $G_N$ upon $Z$, where $Z\in \mathbb{R}^{N\times K}$ is the latent representation given by an encoder network $f_\theta$ by taking the nodes feature $X$, \ie $z_i=f_{\theta}(x_i)$. $K$ indicates the dimension of the representation vectors and is determined by the network $f_{\theta}$. 

In graph $G_N$, the affinity matrix $W\in \mathbb{R}^{N\times N}$ is encoded by the similarities between vertices, which is obtained by:
\begin{equation}
    W_{ij} = \left\{
    \begin{aligned}
       & [z_i^\top z_j]_{+}^ {\beta} , & \text{if} ~ i\neq j \land x_j \in N_k(x_i), \\
       & 0, & \text{otherwise.}
    \end{aligned}
    \right.
\end{equation}
where $N_k(x_i)$ is a similarity-based neighborhood set, \ie the set of $k$ nearest neighbors of $x_i$ in $X$, and $\beta$ is a parameter for diffusion on region manifolds\cite{chandra2016fast,iscen2017efficient}which we simply set as $\beta$ in our experiments.   

Suppose we have clean nodes  along with some noisy nodes, label propagation spreads the label information of each node to the other nodes based on the connectivity in the graph $G_N$. 
To weaken the influence of noisy labels, we set $\tilde{y}_i=y_i$ to the one-hot label vector of $x_i$ if $x_i$ is selected as a clean sample by Eq.\eqref{eq:clean-selection}, otherwise we use category prediction which is a $C$-dimensional vector representing the belongingness of $x_i$ to the $C$ known classes, \ie $\Tilde{y}_i = \hat{p}_i$, as shown in Eq. \eqref{eq:soft-prediciton}. 
The propagation process is repeated until a global equilibrium state is achieved, and each example is assigned to the class from which it has received the most information. 

Formally, for graph $G_N=(W,X)$, $D$ is the degree matrix ( a diagonal matrix with entries $D_{ii}=\sum_{j}W_{ij}$), label propagation \cite{iscen2019label} can be computed by minimizing
\begin{equation}\label{eq:label-propagatio  n}
    J(\bar{Y})= \sum_{i=1}^N \| \bar{y}_i - \tilde{y}_i \| + \alpha \sum_{i,j=1}^N W_{ij}\|\frac{1}{\sqrt{D_{ii}}} \bar{y}_i -\frac{1}{\sqrt{D_{jj}}} \bar{y}_j  \|^2
\end{equation}
where $\alpha$ is a regularization parameter to balance the fitting constraint (the first term) and the smoothing term (the second term). The fitting constraint encourages the classification of each node to their assigned label, and the smoothing term encourages the outputs of nearby points in the graph to be similar \cite{iscen2019label}. The obtained $\bar{Y}=[\bar{y}_1,\ldots, \bar{y}_N] \in \mathbb{R}^{N\times C}$ is the  refined soft pseudo-labels for $X$ after label propagation, further we transform $\bar{Y}$ into hard pseudo-labels by taking the largest prediction score to guide the training. 

Finally, we use a sufficiently high threshold $\eta \in [0,1]$ to select a reliable subset of nodes as the clean dataset:
\begin{equation} \label{eq:clean-selection}
    g_i = \left\{
    \begin{aligned}
        & 1, & \text{if}~ \bar{Y}_{i y_i}^{(t)} > \frac{1}{C}, \\
        & \mathbb{I} [\max_{c} \bar{Y}_{ic}^{(t)} > \eta], & \text{otherwise.}~~~~~ \\
    \end{aligned}
    \right.
\end{equation}
where $\mathbb{I} (\cdot)$ is the indicator function which outputs 1 if the expression holds and 0 otherwise.
$t$ denotes the number of iteration rounds.
$g_i$ is a binary indicator representing the conservation of node $v_i \in V$ when $g_i=1$ and the removal of node $v_i$ when $g_i=0$. 
Thus, the clean node set $V_{cln} = \{v_i\| \forall v_i\in V \wedge g_i =1 \} =  V \backslash \{v_i\| \forall v_i\in V \wedge g_v =0\} $.

\subsection{Open-Set Prototype Learning via Regions}
\textbf{Overview. }
With the filtered training data (clean node set $V_{cln}$), different from the conventional classification paradigm which directly feeds node features into a GNN to predict the class label, we aim at learning multiple representative prototypes for each category, and predicting class label by calculate the similarity between the node and prototypes. 
Upon the latent representation $Z$, in the latent representation space, we learn the prototypes. The prototype pool can be represented as $P=\{P_1, P_2,...,P_C\}$ where $C$ is the number of known classes. 
$P_c = \{p_{c,1}, p_{c,2}, \cdots, p_{c,K_c}\} \in \mathbb{R}^{K_c \times D}$ denotes the prototypes of category $c$ and $K_c$ indicates the number of prototypes of category $c$. 
Given a node $x_i$, after obtaining the high-level feature representation $z_i$, we compare it with all prototypes via calculating cosine similarity:
\begin{equation}
    s_{c,k}^{(i)}=\frac{z_i \cdot p_{c,k}}{\|z_i\| \|p_{c,k}\|}
\end{equation}
After obtain similarity with all prototypes,we regard the class-wise largest similarity score to obtain the scores of  $x_i$ belonging to  class $c$, \ie,
\begin{equation}\label{eq:similarity-score}
    s_c^{(i)}=\max_k(s_{c,1}^{(i)}, \ldots, s_{c,k}^{(i)}, \ldots, s_{c,K_c}^{(i)})
\end{equation}
And the vector of score of belonging between $x_i$ and each class is 
\vspace{2pt}
\begin{equation}\label{eq:soft-prediciton}
   \hat{p}_i = 
   [s_1^{(i)}, s_2^{(i)}, \ldots, s_C^{(i)}]
\end{equation}
The final classification prediction of $x_i$ is 
\begin{equation}\label{eq:classification-prediction}
    \hat{y}_i = \arg\max_c{\hat{p}_{i,c}} = \arg\max_c{s_c^{(i)}}
\end{equation}
\noindent 
\textbf{Open-Set Prototype Learning. }
The open-set node classification is different with the closed set classification problem, in which the class boundaries of known classes require to be tight and clear. Traditional prototype \cite{yang2018robust} learning normally use typical interior prototypes (such as the mean vectors of node representation in the same class) to represent the class. 
However, we argue that border prototypes are also crucial for classification, especially for open-set scenarios, since they provide more detailed information for preserving discrimination between classes and the information of class boundaries. 
We believe the combination of interior prototypes and border prototypes can well relieve the intra-class variety problem caused by the noise and inter-class confusion problem, and obtain tight and clear boundaries for known class, reserve more space for unknown classes. 

Since the original training data containing IND and OOD noises, to avoid the ambiguity problem caused by noise data, we optimize the prototype learning based on the clean nodes $V_{cln}$ and the refined  label annotations $\bar{Y}$. 
We first divide the latent space into regions by clustering the clean nodes where any clustering algorithm could be used, and we use $K$-means for its simplicity. 
With the obtained clusters, under the guidance of label matrix $\bar{Y}$, we use homogeneous clusters to update the interior prototypes and use non-homogeneous clusters to obtain border prototypes. 

To obtain the most representative interior prototype for each class, we make interior prototypes as trainable weights of a feed forward network $f_\phi$ and initialized with He initialization\cite{liu2022graph}. 
For brevity, we denote the interior prototypes by matrix 
\begin{equation}
    P_I = [P^I_{1},\ldots, P^I_C]\in \mathbb{R}^{C \times D}
\end{equation}
$P^I_c$ is the interior prototype for class $c$.
We update the interior prototype network $f_\phi$ and representation learning network $f_\theta$ iteratively. 
Since a rapidly changing of prototypes may disorganize the representativeness of the learned prototypes and make the training process unstable, we utilize a small learning rate to dynamically and smoothly update the prototypes through back-propagation:
\begin{equation}
    P^I_{c,(t)} = P^I_{c,(t-1)} - \varphi \frac{\partial \mathcal{L}}{\partial P^I_c}
\end{equation}

where $P^I_{c,(t)}$ and $P^I_{c,(t-1)}$ are the interior prototype of class $c$ at epoch $t$ and epoch $t-1$ respectively. $\mathcal{L}$ is the total loss shown as Eq. \eqref{eq:total-loss}. $\varphi$ is the learning rate which is set as small numbers. 
Note that here we only use the samples in the homogeneous clusters to update the corresponding interior prototypes, \ie we select the regions containing nodes belonging to a single class, and use these nodes to update the interior representative prototype for the
corresponding class. 

Towards border prototypes, the idea is to analyze those regions which contain nodes belonging to different classes. 
We obtain border prototypes by computing the mean vector of nodes with the same class label in non-homogeneous clusters.
For example, suppose a cluster $R_k$ insists of samples of two classes, \ie $M_k = \bar{V}_1^k \cup \bar{V}_2^k \cup \ldots \cup \bar{V}_C^k$, where $\bar{V}_c^k$ consists of a couple of nodes from class $c$, and there exists  $ m,n \in \{1,\ldots, C\}$ such that $\bar{V}_m^k\neq \emptyset ~\wedge ~\bar{V}_n^k \neq \emptyset$, then we can obtain border prototypes of class $m$ and $n$ by:
\begin{equation}
    P_m^k = \frac{1}{\|\bar{V}_m^k\|} \sum_{v_i \in \bar{V}_m^k} z_i, ~~
    P_n^k = \frac{1}{\|\bar{V}_n^k\|} \sum_{v_i \in \bar{V}_n^k} z_i
\end{equation}
It is the same for clusters that contain nodes from several different classes. 

We utilize cross-entropy loss to train the encoder network $f_\theta$ and prototype network $f_\phi$ on clean nodes $V_{cln}$, under the supervision of refined class label $\bar{Y}$:

\begin{equation}
    \mathcal{L}_{cls} = -\frac{1}{N} \sum_{i=1}^N \bar{y}_i \log\frac{exp(s_{\bar{y}_i}^{(i)}/T)}{\sum_{c=1}^C \exp{(s_c^{(i)} / T)}} 
\end{equation}
where $T$ is a temperature hyperparameter that we introduced to make the results more differentiated \cite{agarwala2020temperature}. 

With the adoption of both interior and border prototypes, the diversity of prototypes within a class can be well mined.
To further relieve the inter-class confusion problem, we hope the prototypes of different categories also away from each other. 
Thus, we enhance the diversity of the prototypes of known classes by adopting the orthogonal constraint to keep the orthogonality of interior prototypes by using the diversity loss:
\begin{equation}
    \mathcal{L}_{div} = \|P_I P_I^\top -I\|_F^2
\end{equation}
where $\|\cdot\|_F$ is the Frobenius-norm and $I$ is the identity matrix of any desired dimension. 
The overall loss function of ROG$_{PL}$ is:

\begin{equation}\label{eq:total-loss}
    \mathcal{L} = \mathcal{L}_{cls} + \lambda \mathcal{L}_{div}
\end{equation}
where $\lambda$ is the loss hyper-parameter.

So far, we obtain  interior prototype and  border prototypes for each known class, and we predict the class label of clean nodes through Eq. \eqref{eq:similarity-score} and \eqref{eq:classification-prediction}. 
To make a hard prediction, we adopt a probability threshold $\tau$, such that a testing point $x_i$  is deemed as an unknown class sample  if 
$\max_{c}\hat{p}_{i,c} < \tau$.
 
\section{Experiments} 
We design our experiments to evaluate ROG$_{PL}$, focusing on the following aspects: \textsl{open-set classification comparison}, \textsl{robustness analysis}, and \textsl{ablation study}.
Codes will be available online. 

\subsection{Experimental Setup}
\textbf{Dataset and Metrics. }
To evaluate the performance of the proposed framework for robust open-set node classification, 
We conducted experiments on three main benchmark graph datasets~\cite{wu2020openwgl,zhu2022shift}, namely 
Cora\footnote{https://graphsandnetworks.com/the-cora-dataset/}, Citeseer\footnote{https://networkrepository.com/citeseer.php} \cite{yang2016revisiting}, and Coauthor-CS\footnote{https://docs.dgl.ai/en/0.8.x/generated/dgl.data.Coauthor-\\CSDataset.html}~\cite{zhou2023robust},
which are widely used citation network datasets.
The statistics of the datasets are presented in the Appendix.
In terms of the metrics, following the study of \textit{Wu} \etal \cite{wu2021ngc}, we adopt macro-F1 and AUROC to measure the performance.

\textbf{Implementation Details. } 
In the experiments, we adopt GCN~\cite{kipf2016semi-han10} as backbone neural network for the encoder $f_\theta$, configured with two hidden layers with a dimension of 128. The prototype network  $f_\phi$ configured with a linear layer with a dimension of $C+1$.
ROG$_{PL}$ is implemented with PyTorch and the networks are optimized using adaptive moment estimation with a learning rate of $10^{-3}$. The balance parameters $\lambda$ is set to $10^{-2}$. The threshold $\eta$ were selected by a grid search in the range from $0$ to $1$ with a step of $10^{-1}$. The number of nearest neighbors $k$ were selected by a grid search in the range from $30$ to $35$ with a step of $1$.

For each experiment, the baselines and the proposed method were applied on the same training, validation, and testing datasets.
All the experiments were conducted on a workstation equipped with an Intel(R) Xeon(R) Gold 6226R CPU and an Nvidia A100 GPU.

\textbf{Test Settings. }
To evaluate the performance of open-set node classification, for each dataset, the data of several classes were held out as the unknown classes for testing and the remaining classes were considered as the known classes. 70\% of the known class nodes were sampled for training, 10\% for validation, and 20\% for testing. 

To assess the performance of the proposed ROG$_{PL}$ framework on graph data with different noises, we tested it with mixed IND and OOD noise.
In the training set, we randomly selected 5\%/25\%/50\% of the known class samples to be IND noise, and randomly replaced their ground-truth labels with wrong known class labels. 
Then, we used the nodes from neither the known classes of the training set nor the unknown classes of the test set as OOD noise, and randomly assigned  known class labels to them.
The samples with right known class labels (clean data), the known class samples with incorrect labels (IND noise) and the far-away unknown class samples with wrong known class labels (OOD noise) constitute the final training set. 
The test set includes the known class samples and the unknown class samples, where the unknown classes are different from the classes of OOD noise.
The setting of inductive learning was adopted for all the experiments, where no information about the unknown class in the test set (such as the feature $x_i$  or other side information of unknown classes) is utilized during training or evaluation.

\begin{table*}[htb] 
\centering
\setlength{\tabcolsep}{0.8mm}{
\begin{tabular}{|c|cc||cccccccccc||c|}
\hline  
\multicolumn{3}{|c||}{Methods} & {\small GCN}{\scriptsize \_soft} & {\small GCN}{\scriptsize \_sig} & {\small GCN}{\scriptsize \_soft$\_\tau$} & {\small GCN}{\scriptsize \_sig$\_\tau$}  & {\small NWR\_$\tau$   } & {\small  Openmax} & {\small OpenWGL }  & {\small $\mathcal{G}^2Pxy$ } & $~~~$NGC$~~~$    & $~~~$PNP$~~~$    & ROG$_{PL}$	     \\
\hline
\hline
\multirow{6}{*}{\rotatebox{90}{Cora}} & \multirow{2}{*}{5\%}   & F1  &	63.73	&	64.12	&	70.53	&	57.80		&	69.41	&	60.40	&	\underline{73.03}	&	72.00	&	59.69	&	70.26		&	\textbf{78.36}	\\

 &                        & {\scriptsize AUROC} &	fail	&	fail	&	\underline{82.82}	&	81.40		&	75.05	&	67.26	&	79.36	&	78.30	&	78.20	&	82.04		&	\textbf{91.00}	\\

\cline{2-3}
& \multirow{2}{*}{25\%}  & F1    &	64.31	&	63.78	&	\underline{70.69}	&	66.91		&	68.98	&	59.82	&	70.58	&	69.92	&	52.13	&	64.78		&	\textbf{77.59}	\\
        &                         & {\scriptsize AUROC} &	fail	&	fail	&	77.92	&	79.94		&	81.69	&	\underline{87.61}	&	77.90	&	77.36	&	80.22	&	84.87		&	\textbf{91.59}	\\  

\cline{2-3}
                      & \multirow{2}{*}{50\%}  & F1    &	60.96	&	59.73	&	62.53	&	62.93		&	61.93	&	53.50	&	63.59	&	57.52	&	62.70	&	\underline{66.29}		&	\textbf{76.90}	\\

                      &                         & {\scriptsize AUROC} &	fail	&	fail	&	79.46	&	78.52		&	78.73	&	80.88	&	\underline{85.40}	&	63.68	&	83.64	&	83.54		&	\textbf{90.47}	\\  
\hline   \hline 
\multirow{6}{*}{\rotatebox{90}{Citeseer}} & \multirow{2}{*}{5\%}   & F1    &	39.03	&	37.74	&	59.13	&	59.93		&	52.28	&	33.23	&	59.87	&	52.40	&	59.44	&	\underline{59.99}		&	\textbf{62.24}	\\
                      &                        & {\scriptsize AUROC} &	fail	&	fail	&	85.67	&	85.56		&	76.13	&	55.01	&	\underline{86.08}	&	67.75	&	84.05	&	85.70		&	\textbf{87.84}	\\  
\cline{2-3}
                      & \multirow{2}{*}{25\%}  & F1    &	38.30	&	38.02	&	23.04	&	57.79		&	48.51	&	36.70	&	57.42	&	53.53	&	55.42	&	\underline{59.83}		&	\textbf{63.89}	\\
                      &                         & {\scriptsize AUROC} &	fail	&	fail	&	\underline{84.50}	&	82.26		&	72.59	&	75.68	&	82.34	&	53.97	&	80.46	&	84.03		&	\textbf{88.84}	\\  
\cline{2-3}
                      & \multirow{2}{*}{50\%}  & F1    &	30.39	&	31.71	&	35.41	&	35.54		&	35.65	&	24.83	&	40.46	&	41.92	&	\underline{56.39}	&	51.81		&	\textbf{57.08}	\\
                      &                         & {\scriptsize AUROC} &	fail	&	fail	&	64.20	&	50.74		&	61.22	&	67.89	&	79.54	&	38.03	&	77.43	&	\underline{80.40}		&	\textbf{85.64}	\\  
\hline \hline
\multirow{6}{*}{\rotatebox{90}{ Coauthor-CS}} & \multirow{2}{*}{5\%}   & F1    &	75.44	&	76.50	&	76.63	&	71.70		&	75.43	&	62.68	&	74.09	&	\underline{83.45}	&	77.61	&	\textbf{83.56}		&	81.68	\\
                      &                        & {\scriptsize AUROC} &	fail	&	fail	&	83.14	&	86.61		&	\underline{89.57}	&	81.65	&	87.39	&	82.95	&	85.57	&	85.55		&	\textbf{93.25}	\\  
\cline{2-3}
                      & \multirow{2}{*}{25\%}  & F1   &	74.58	&	75.19	&	\underline{81.92}	&	73.38		&	65.64	&	66.38	&	72.98	&	67.21	&	69.18	&	79.37		&	\textbf{83.32}	\\
                      &                         & {\scriptsize AUROC} &	fail	&	fail	&	87.77	&	88.35		&	\underline{88.69}	&	80.62	&	86.35	&	86.03	&	85.11	&	85.47		&	\textbf{94.06}	\\  
\cline{2-3}
                      & \multirow{2}{*}{50\%}  & F1   &	72.93	&	60.77	&	68.87	&	52.63		&	\underline{78.66}	&	59.97	&	54.70	&	72.96	&	63.06	&	68.40		&	\textbf{79.29}	\\
                      &                         & {\scriptsize AUROC} &	fail	&	fail	&	75.02	&	83.97		&	\underline{87.58}	&	81.48	&	85.13	&	72.87	&	84.27	&	84.85		&	\textbf{93.31}	\\

\hline
\end{tabular}}
\caption{Comparison of open-set node classification in test F1-score and AUROC (\%) on three datasets, where IND noise (5\%/ 25\%/ 50\%) and near OOD noise is injected into training set.} 
\label{tb:near-ood}
\end{table*}

\textbf{Baselines. }
We compare ROG$_{PL}$ with 10 baselines, which are from three categories. 
\begin{itemize}[leftmargin=*]
    \item 1) Closed-set classification methods: GCN\_soft and GCN\_sig. They are GCNs \cite{kipf2016semi-han10} with a softmax layer or a multiple 1-vs-rest of sigmoids layer as output layer. 
    \item 2) Open-set classification methods: GCN\_soft\_$\tau$, GCN\_sig\_$\tau$, NWR\_$\tau$~\cite{tang2022graph},
    Openmax~\cite{bendale2016towards-holder3-han2}, OpenWGL~\cite{wu2020openwgl} and $\mathcal{G}^2Pxy$~\cite{zhang2023g2pxy}. Specifically, GCN\_soft\_$\tau$, GCN\_sig\_$\tau$ and NWR\_$\tau$ are  GCN\_soft, GCN\_sig and the original NWR~\cite{tang2022graph} methods by added a threshold chosen from $\{0.1, 0.2,\ldots,0.9\}$  to perform open-set recognition.
    Openmax~\cite{bendale2016towards-holder3-han2} is an open-set recognition model based on ``activation vectors'' (i.e. penultimate layer of the network). 
OpenWGL~\cite{wu2020openwgl} and $\mathcal{G}^2Pxy$~\cite{zhang2023g2pxy} are two open-set node classification methods for graph data, which has no robust learning ability. 

\item 3) Robust open-set classification methods: NGC~\cite{wu2021ngc} and PNP~\cite{sun2022pnp}.
NGC~\cite{wu2021ngc} is an open-world noisy data learning method for image classification which employs geometric structure and model predictive confidence to collect clean sample.
PNP\_$\tau$~\cite{sun2022pnp} is  a robust classifier learning method for image data with IND and OOD noise, where data augmentation is used to help the identification of noisy samples.
To perform open-set recognition, we adopt a  threshold chosen from $\{0.1, 0.2,\ldots,0.9\}$ to PNP.
\end{itemize}

Note that the graph data are first embedded by a GCN before being feed into the models that cannot handle graph data.
A detailed introduction can be found in the Appendix. 

\subsection{Open-Set Node Classification with Complex Noisy Graph Data}
Considering that real-world scenarios are complex and noisy data vary across different tasks, we assessed the proposed model for open-set classification with IND noise and two types of OOD data: near OOD data and far OOD data. 
Here, OOD data include OOD noise in the training set and out-of-distribution samples of unknown classes that occur during testing. 

\subsubsection{Open-Set Classification with IND Noise and Near OOD Data. }
In this experiment, for each dataset, following the setting of OpenWGL \cite{wu2020openwgl}, the data of the last class were held out as the unknown class for testing, and the data of the second last class which were re-assigned with random known class labels were set as the OOD noise data and injected into the training set. The remaining classes were considered as known classes, while the known class samples in training set were re-assigned with wrong known class labels with a rate of 5\%, 25\% and 50\% (\ie IND noise rate), respectively. 

Table \ref{tb:near-ood}  lists the 
macro-F1 and AUROC scores for open-set node classification 
with  near OOD noise and different proportions of IND noise. 
It is observed that ROG$_{PL}$ generally obtain the best performance on the benchmarks. 
This shows that ROG$_{PL}$  can better distinguish between a known class and an unknown class, even though there is a large amount of complex and diverse noises  during training. 
Specifically, ROG$_{PL}$ achieves an average of 6.23\% improvement over the second-best method (PNP) in terms of F1 score and an average of 6.62\% improvement in terms of AUROC on the three datasets. 

Furthermore, we examined the detailed classification accuracy in  terms of known classes and unknown classes.
We found that to gain the ability of unknown class detection, compared to the closed-set classifier, there is a slight decrease in the performance of known class classification, \ie, from 86.33\% (GCN\_soft) to 75.48\% (ROG$_{PL}$) on average, while the unknown class detection accuracy is increased from 0\% to a remarkable 83.78\% on average.  Moreover, compared to other open-set node classification methods, such as $\mathcal{G}^2Pxy$, ROG$_{PL}$ achieves an average improvement of 11.9\%, 13.29\% and 9.94\% in accuracy of known classification, unknown detection and overall classification, respectively. The average improvement in F1 score is 9.94\%.
Details are provided in the Appendix. 

\begin{table*}[htb] 
\label{tb:far-ood}
\centering
\setlength{\tabcolsep}{0.8mm}{  
\begin{tabular}{|l||cc|cc|cc||cc|cc|cc|}
\hline
\multirow{3}{*}{Methods} & \multicolumn{6}{c||}{5\% far OOD noise}   & \multicolumn{6}{c|}{25\% far OOD noise}   \\  \cline{2-13}
                       & \multicolumn{2}{c|}{Cora}     & \multicolumn{2}{c|}{Citeseer} & \multicolumn{2}{c||}{Coauthor-CS}  
                       & \multicolumn{2}{c|}{Cora}     & \multicolumn{2}{c|}{Citeseer} & \multicolumn{2}{c|}{Coauthor-CS}   \\ \cline{2-13} 
                       & F1      & {\scriptsize AUROC}               & F1       & {\scriptsize AUROC}              & F1    & {\scriptsize AUROC}            & F1      & {\scriptsize AUROC}               & F1       & {\scriptsize AUROC}             & F1    & {\scriptsize AUROC}      
                       \\ \hline \hline
                   
GCN\_soft	        & 64.67	&	fail	&	39.04	&	fail	&	60.36	&	fail	&	63.12	&	fail	&	39.04	&	fail	&	77.08	&	fail	\\ 
GCN\_sig	        & 64.11	&	 fail	&	41.33	&	fail	&	69.32	&	fail	&	63.44	&	fail	&	40.78	&	fail	&	65.94	&	fail\\
GCN\_soft$\_\tau$   & 71.81	&	74.61	&	58.57	&	82.35	&	65.81	&	78.37	&	70.38	&	75.13	&	56.09	&	81.61	&	60.48	&	72.37 \\
GCN\_sig$\_\tau$    & 62.65	&	77.54	&	54.60	&	\underline{84.58}	&	54.23	&	86.73	&	69.23	&	76.38	&	57.41	&	81.81	&	57.75	&	84.31 \\
NWR\_$\tau$        &71.94	&	76.58	&	48.35	&	75.06	&	\underline{79.46}	&	\underline{87.65}	&	68.79	&	80.67	&	48.08	&	71.67	&	54.77	&	83.76\\
Openmax         	& 28.37	&	79.06	&	26.64	&	81.17	&	66.90	&	82.10	&	31.41	&	\underline{84.40}	&	15.55	&	79.03	&	67.23	&	85.07\\
OpenWGL             &71.83	&	76.76	&	58.51	&	\textbf{84.78}	&	76.38	&	87.24	&	67.87	&	74.50	&	57.28	&	\underline{82.75}	&	75.71	&	\underline{85.10}\\
$\mathcal{G}^2Pxy$    &69.24	&	 \underline{80.67}	&	43.26	&	62.53	&	66.98	&	69.57	&	66.54	&	75.26	&	41.67	&	60.95	&	70.45	&	68.34\\
NGC	                & 57.58	&	78.57	&	52.30	&	78.07	&	66.96	&	84.47	&	59.86	&	76.88	&	55.41	&	81.22	&	64.84	&	83.96\\
PNP         & \underline{72.21}	&	79.51	&	\underline{60.22}	&	83.93	&	77.50	&	87.42	&	\underline{72.19}	&	77.47	&	\underline{57.58}	&	81.94	&	\underline{81.34}	&	84.23\\
\hline
\hline
ROG$_{PL}$	            & \textbf{77.62}	&	\textbf{87.89}	&	\textbf{60.90}	&	84.50	&	\textbf{83.11}	&	\textbf{92.34}	&	\textbf{76.31}	&	\textbf{86.48}	&	\textbf{58.08}	&	\textbf{85.49}	&	\textbf{82.06}	&	\textbf{89.78}\\

\hline
\end{tabular}}
\caption{Comparison of open-set node classification in test F1-score and AUROC (\%) on three datasets, where IND noise (5\%) and far OOD noise (5\% / 25\%) from the Pubmed dataset is injected into the training set.}
\end{table*}

\subsubsection{Open-Set Node Classification with IND Noise and Far OOD Data. }
To investigate the effect of rough OOD noise, we used OOD samples that are from different dataset as the source of OOD noise. Specifically, we randomly selected samples of the first two classes from Pubmed \cite{hu2020open} and mixed them with data from Cora, Citeseer and Coauthor-CS to create the training set with far OOD noise, where the OOD noise rate is set to 5\% or 25\%. 
Each node from Pubmed was randomly assigned to a known class label, and edges between the node and its $k$-nearest neighbors were added, where $k$ is a random integer in the range from 1 to 5. 
We also added some samples from the remaining categories in the Pubmed dataset to the test set in corresponding proportion, to evaluate the performance of ROG$_{PL}$ on the samples from far open-set.
The other settings are the same as the experiment of near OOD data, and the IND noise rate is set as 5\%. 

The results of open-set node classification with far OOD noise are presented in Table 2. 
The results show that ROG$_{PL}$ generally outperforms the baselines, achieving an average improved of 2.84\% in F1 and an average improvement of 5.33\% in AUROC, compared to the global second-best method PNP. 
The detailed classification accuracy in terms of known classes and unknown classes with far OOD noise are given in the Appendix. 

\subsection{Robustness Analysis under Different Noise Rate}
In this section, we evaluate the robustness of ROG$_{PL}$ for different levels of IND and OOD noise for the open-set node classification task.
We first examine how ROG$_{PL}$ reacts to different IND noise rate. 
We kept the OOD noise rate constant and used the same setting as the experiment of Table \ref{tb:near-ood}, while we varied the IND noise rate from 0\%, 5\%, 25\%, 50\% to 75\%. Fig. 2(a) and Fig. 2(b) show the results on the Cora and Citeseer datasets, respectively. 
We observe that ROG$_{PL}$ maintains a relatively stable performance when the noise rate is within a certain range, for example no more than 50\%.  
However, once the IND noise exceeds a certain threshold, ROG$_{PL}$'s performance drops sharply.  

Additionally, we investigate the performance of ROG$_{PL}$ in terms of different OOD noise rate, by keeping the IND noise rate constant as 5\%, while varying the far OOD noise rate from 0\%, 5\%, 25\%, 50\% to 75\% . 
The results on Cora and Citeseer are shown in Fig.2(c) and Fig. 2(d)respectively.
It can be observed that ROG$_{PL}$ maintains a surprisingly stable performance even with large amounts of far OOD noise in the training data, which demonstrates the strong robustness of ROG$_{PL}$ against OOD noise. 

\begin{figure*}[htbp]
    \centering
    \subfigure[Cora]{
        \includegraphics[width=0.23\linewidth]{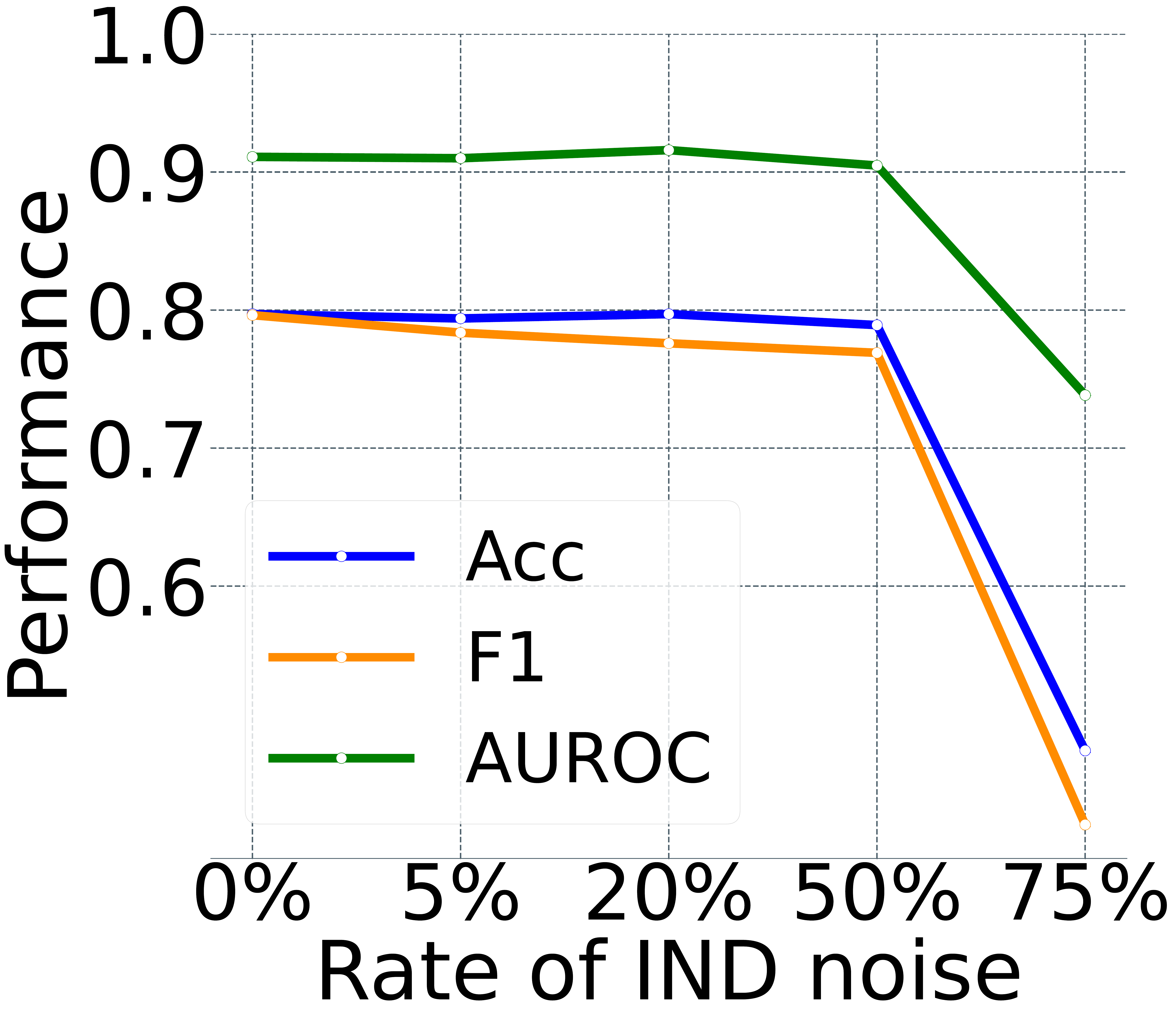}
        \label{fig:cite-cora-ind-noise-rate}
    }
    \subfigure[Citeseer]{
	\includegraphics[width=0.23\linewidth]{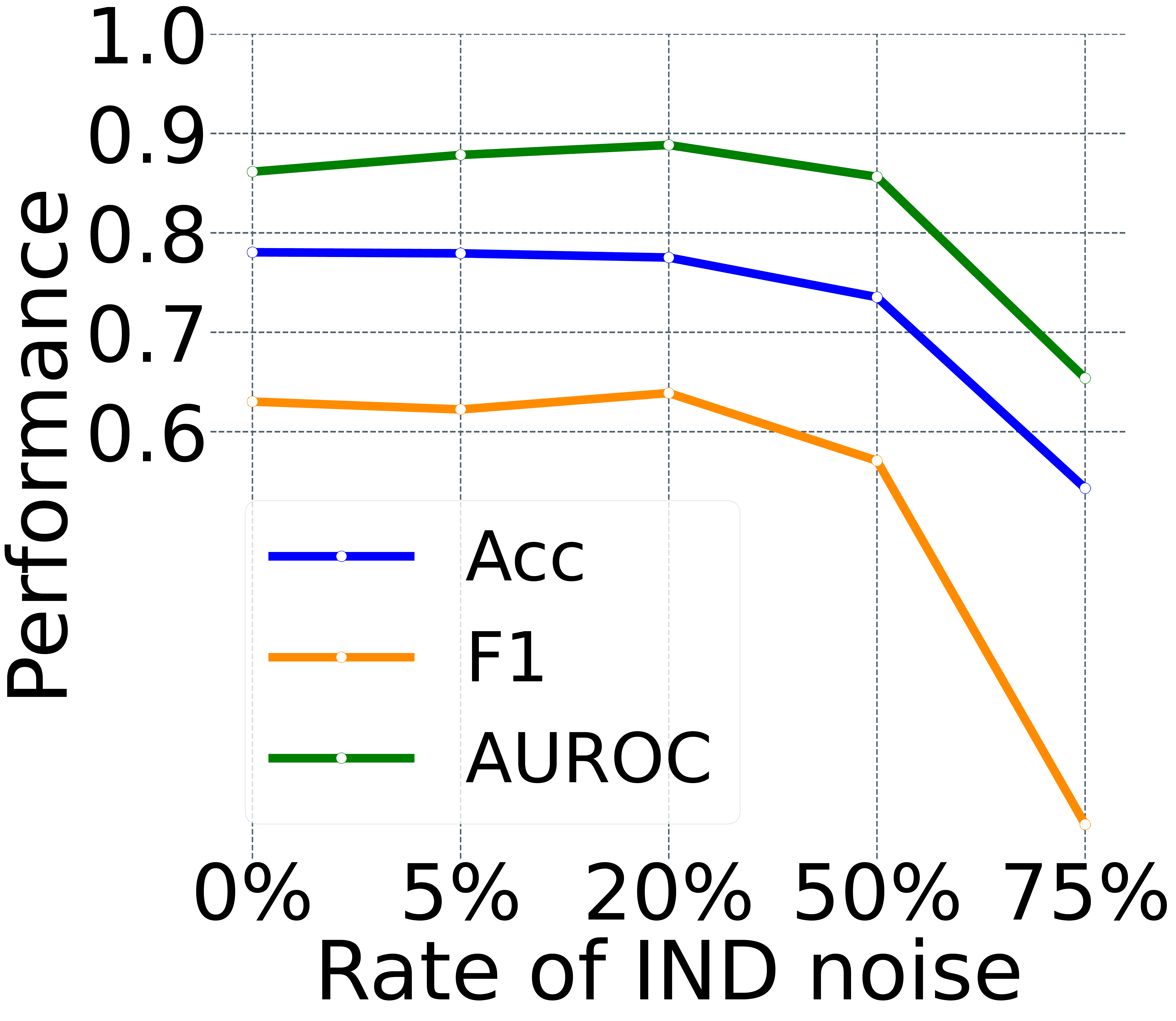}
        \label{fig:cite-citeseer-ind-noise-rate}
    }
        \subfigure[Cora]{
        \includegraphics[width=0.23\linewidth]{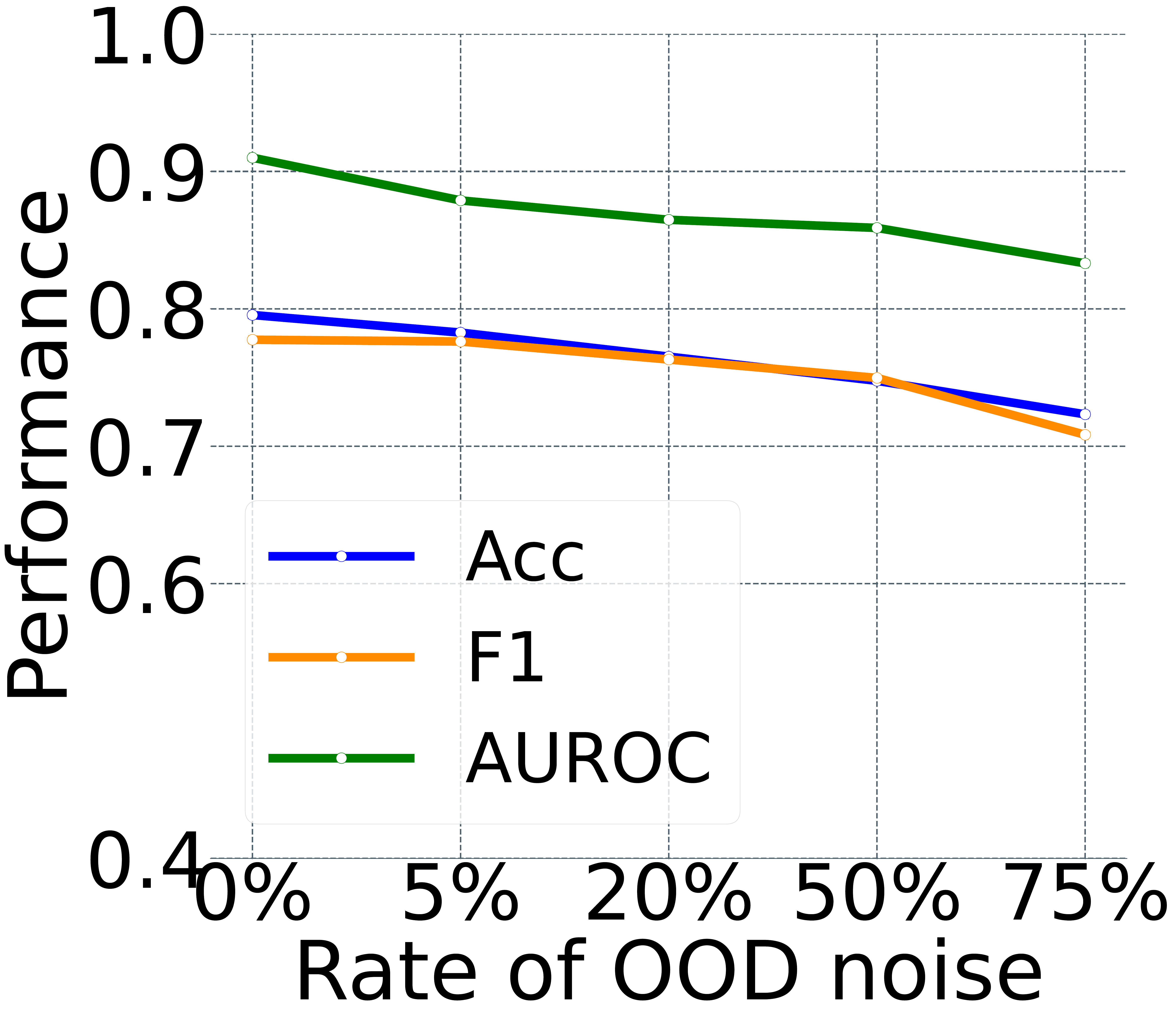}
        \label{fig:cite-cora-ood-noise-rate}}
    \subfigure[Citeseer]{
	\includegraphics[width=0.23\linewidth]{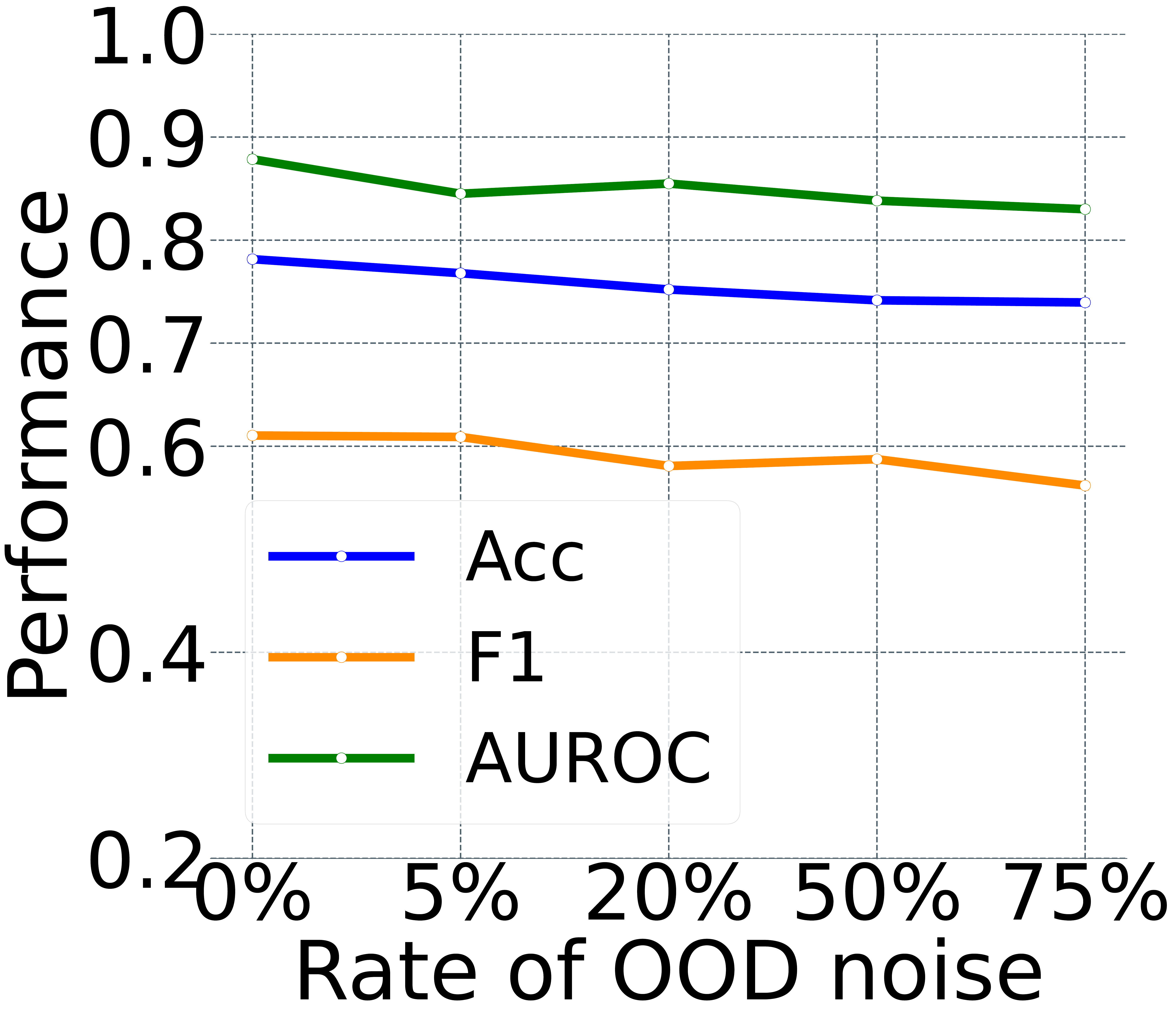}
        \label{fig:cite-citeseer-ood-noise-rate}
    }
    \caption{The performance of ROG$_{PL}$ with respect to different IND noise rate and OOD noise rate on Cora and  Citeseer datasets. }
    \label{fig:noise-rate}
\end{figure*}

\begin{table}[htb]
\centering
\setlength{\tabcolsep}{0.2mm}{
\begin{tabular}{|l|cc|cc|cc|}
\hline
\multirow{2}{*}{Methods} & \multicolumn{2}{c|}{Cora}     & \multicolumn{2}{c|}{Citeseer} & \multicolumn{2}{c|}{Coauthor-CS}      \\ \cline{2-7} 
                                   & F1      & { \small AUROC}                & F1       & {\small AUROC}               & F1    & {\small AUROC}            \\ \hline \hline
                                   
{\small ROG$_{PL}\neg G_N$}	& 74.87	& 89.90	& 61.79	& 86.94	& 81.08	& 91.10\\
{\small 
ROG$_{PL}\neg$denoise } 	&	77.24	&	86.56  	&   61.24 	&	86.99  	&	81.03 	&	87.86 	\\
{\small ROG$_{PL}\neg\mathcal{L}_{div}$} 	&	73.67 	&	89.85  	&  60.35 	&	86.73  	&	79.88 	&	90.32 	\\
{\small ROG$_{PL}\neg$region}	& 76.54	& 84.11		& 61.63	& 87.49	& 80.89	& 86.45\\
{\small ROG$_{PL}$}  	& \textbf{78.36}	& \textbf{91.00}	& \textbf{62.24}	& \textbf{87.84}	& \textbf{81.68}	& \textbf{93.25} \\
\hline
\end{tabular}}
\caption{Ablation study of robust open-set node classification in test F1 score and AUROC (\%) on three datasets, 
where IND noise (5\%) and near OOD noise is injected into the training set.}
\label{tb:ablation-study}
\end{table}

\subsection{Ablation Study }
We compare variants of ROG$_{PL}$ in an ablation study to evaluate the effect of its main modules and settings:
\begin{itemize}

    \item {} ROG$_{PL}\neg G_N$: a variant of ROG$_{PL}$ without building $k$-nearest neighbor graph $G_N$, and original graph $G$ is used in the  label propagation stage.
    \item {} ROG$_{PL}\neg$denoise: a variant of ROG$_{PL}$ without the label propagation based denoising module. 
    \item {} ROG$_{PL}\neg$region: a variant of ROG$_{PL}$ without clustering, \ie the interior prototype of each class is updated by the clean nodes from all classes, and there is no border prototypes. 
    \item {} ROG$_{PL}\neg\mathcal{L}_{div}$: a variant of ROG$_{PL}$ with loss $\mathcal{L}_{div}$ removed. We only utilize classification loss $\mathcal{L}_{cls}$ to train the encoder network $f_\theta$ and prototype network $f_\phi$. 

\end{itemize}

The performance of the proposed method and its four variants are presented in Table \ref{tb:ablation-study}. 
The results demonstrate that both label propagation based denoising module and region-based prototype learning module are important and building $k$nn graph $G_N$ is necessary. The large gap of performance between \textit{ ROG$_{PL}$} and \textit{ ROG$_{PL}\neg$region } verifies the contribution of open-set prototypes on open-set node classification.

\section{Conclusion}
This paper introduced a novel prototype learning based robust open-set node classification method to learn an open-set classifier from graphs with mixed IND and OOD noisy nodes. 
By correcting noisy labels through similarity-based label propagation and removing low-confidence samples, the proposed method relieve the intra-class variety caused by noise. 
Further, by learning open-set prototypes via non-overlapped regions and  remaining both interior and border prototypes, the method can remedy the inter-class confusion problem, and save more space for open-set classes.
To the best of our knowledge, the proposed ROG$_{PL}$ is the first robust open-set node classification method for graph data with complex noise. 
Experimental evaluations on several benchmark graph datasets demonstrates its good performance.

\section*{Acknowledgments}
This research was supported by National Natural Science Foundation of China (62206179, 92270122),
Guangdong Provincial Natural Science Foundation (2022A1515010129, 2023A1515012584), 
University stability support program of Shenzhen (20220811121315001), 
Shenzhen Research Foundation for Basic Research, China (JCYJ20210324093000002).

\bibliographystyle{aaai24} 
\bibliography{aaai24}
\end{document}